\documentclass[11pt]{article}

\usepackage[preprint]{acl}

\usepackage{amssymb}
\usepackage{amsmath}
\usepackage{amsthm}
\usepackage{times}
\usepackage{latexsym}

\usepackage[T1]{fontenc}

\usepackage[utf8]{inputenc}

\usepackage{microtype}

\usepackage{inconsolata}

\usepackage{graphicx}
\usepackage{booktabs}
\usepackage{multirow}
\usepackage{tikz}
\usetikzlibrary{arrows.meta,positioning}
\usepackage[table]{xcolor}
\usepackage{colortbl}
\theoremstyle{definition}

\definecolor{hlecyan}{HTML}{7DF9FF}    
\definecolor{hlcyan}{HTML}{D6F0FF}     
\definecolor{hlelilac}{HTML}{D8B4FE}   
\definecolor{hllilac}{HTML}{E5D8F2}    

\title{Hubness, Not Anisotropy, Drives Cross-Lingual Retrieval Asymmetry in Multilingual Embedding Models}

\author{\bf Adib Sakhawat, 
{\bf Fardeen Sadab,}
{\bf Atik Shahriar,}\\
Department of Computer Science and Engineering\\
Islamic University of Technology, Dhaka, Bangladesh\\
\texttt{\small\{adibsakhawat, fardeensadab, atikshahriar\}@iut-dhaka.edu}\\
}

\begin{document}
\maketitle
\begin{abstract}
Multilingual embedding models are deployed under the assumption that cross-lingual retrieval is symmetric: if a query in language $A$ retrieves its translation in language $B$, the reverse should also hold. In practice it does not. Using a parallel corpus of $6{,}518$ idiomatic and proverbial expressions in English, Bangla, Hindi, and Arabic, embedded by five production-grade encoders (\textsc{Gemini}, \textsc{Mistral}, \textsc{OpenAI-L}, \textsc{OpenAI-S}, \textsc{Qwen}), we formalise this failure as a deficit in mutual nearest-neighbour reciprocity and test a single mechanistic claim: among the geometric pathologies of multilingual spaces, \emph{hubness} not anisotropy, centroid drift, or magnitude is the dominant causal driver. Across five pre-registered experiments with falsification conditions specified in advance, hub mass dominates a joint regression on reciprocity (49.5\% dominance share, $1.68\times$ the next predictor; partial $R^{2}=0.302$ versus $0.003$ for anisotropy), while a hub-aware score correction (CSLS) closes $63.5\%$ of the worst-to-best reciprocity gap and yields a mean within-model effect size $130\times$ larger than surgical hub-vector ablation. The latter contrast pinpoints the mechanism: hubness is a pathology of the \emph{similarity metric}, not of individual hub vectors. We resolve the well-known anisotropy--hubness paradox by showing the two are statistically dissociable, and we recommend replacing cosine similarity with CSLS as the default retrieval metric for multilingual embedding pipelines.
\end{abstract}

\section{Introduction}
\label{sec:intro}

\begin{figure}[t]
\centering
\resizebox{\columnwidth}{!}{%
\includegraphics{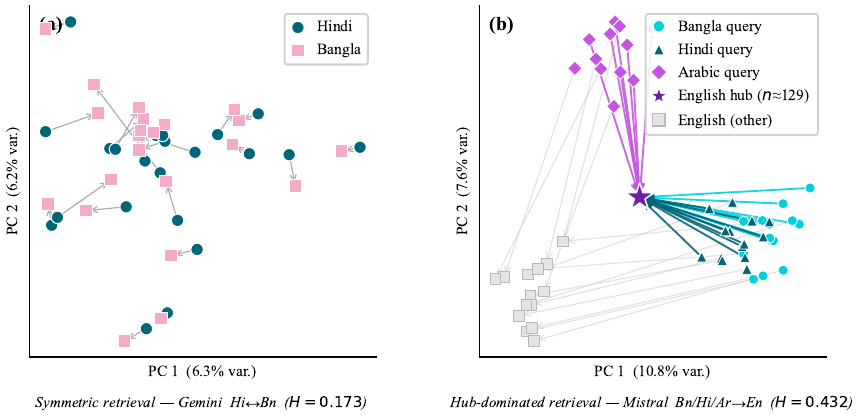}%
}
\caption{Conceptual illustration of hubness in cross-lingual retrieval: a small number of target-language vectors become nearest neighbours for many source-language queries, producing asymmetric retrieval failures.}
\label{fig:hubness_conceptual}
\end{figure}

Multilingual embedding models \textsc{Gemini}, \textsc{Mistral}, the \textsc{OpenAI} \textsc{text-embedding-3} family, \textsc{Qwen} are deployed as drop-in components of cross-lingual retrieval and multilingual RAG on a geometric promise: semantically equivalent texts in different languages should land near one another in a shared space. Retrieval over those spaces is nonetheless conspicuously asymmetric across model families, scripts, and thresholds. Figure~\ref{fig:hubness_conceptual} contrasts the two regimes: a well-behaved space yields symmetric one-to-one alignments (panel~a), while a pathological one shows severe \emph{hubness} dozens of queries from multiple source languages collapsing onto a single popular target (panel~b).

A growing literature catalogues geometric pathologies that plausibly explain such failures anisotropy \cite{ethayarajh2019contextual,mu2017allbuttop}, hubness \cite{radovanovic2010hubs,dinu2014hubness,lazaridou2015hubness}, centroid drift, magnitude variance but reports them as a heterogeneous list without a falsifiable claim about which is causally responsible. We close that gap with a single mechanistic claim, the \emph{Hub-Mediation Hypothesis} ($H_{0}$):

\begin{quote}
\itshape Among the geometric pathologies of multilingual embedding spaces, hubness is the dominant causal driver of cross-lingual retrieval asymmetry. Interventions that suppress hub influence at the score level should restore reciprocity in proportion to the hub mass they neutralise.
\end{quote}

The outcome variable is \emph{retrieval reciprocity} $R$, the mutual nearest-neighbour rate, on which bi-directional retrieval and multilingual RAG actually depend. We test $H_{0}$ on a parallel corpus of $6{,}518$ idiomatic and proverbial expressions in English, Bangla, Hindi, and Arabic embedded by five production encoders ($20$ model$\times$pair observations).

\paragraph{Contributions.} We deliver a causal decomposition of multilingual geometric pathologies in which hubness alone takes $49.5\%$ of explained variance via dominance analysis (anisotropy's partial $R^{2}=0.003$); an experimental dissociation between hub \emph{vectors} and hub \emph{score influence} surgical top-$100$ ablation is inert (Cohen's $d\!\approx\!0.03$) while CSLS \cite{lample2018wordtranslation} reaches $d\!\approx\!2.4$, a $130\times$ ratio that pinpoints the mechanism as score distortion rather than the physical presence of hubs; a practical recommendation, namely that CSLS closes $63.5\%$ of the worst-to-best reciprocity gap without retraining and improves Recall@1 across all five encoders; and a resolution of the anisotropy--hubness paradox, since high-anisotropy/low-hubness (\textsc{Qwen}) and moderate-anisotropy/high-hubness (\textsc{OpenAI-S}) models behave exactly as $H_{0}$ predicts once the two pathologies are statistically separated.

\section{Related Work}
\label{sec:related}

\paragraph{Hubness as a geometric pathology.} \citet{radovanovic2010hubs} introduced hubness as an inherent consequence of the curse of dimensionality, showing that in high-dimensional spaces a small set of points become disproportionately popular nearest neighbours. Subsequent work demonstrated that distance scaling can substantially suppress hubness \cite{schnitzer2012scaling,flexer2015lp}.

\paragraph{Hubness in cross-lingual and zero-shot settings.} \citet{dinu2014hubness} and \citet{lazaridou2015hubness} were the first to demonstrate that cross-space mapping in zero-shot learning including cross-lingual word translation suffers from severe hubness, with hub-aware corrections improving retrieval. \citet{smith2017invertedsoftmax} proposed the inverted softmax as an alternative hubness-mitigating retrieval criterion.

\paragraph{Cross-lingual word embeddings and CSLS.} \citet{lample2018wordtranslation} introduced Cross-domain Similarity Local Scaling (CSLS) as a hub-aware retrieval metric for unsupervised bilingual lexicon induction. \citet{artetxe2018unsupervised} and \citet{joulin2018loss} approached the same problem through better mapping or retrieval-aligned objectives. \citet{glavas2019evaluation} systematised evaluation of cross-lingual word embeddings, and \citet{glavas2020instamap} argued that non-isomorphism, not hubness, is the principal obstacle an alternative hypothesis our experiments speak to directly.

\paragraph{Multilingual sentence embeddings.} The encoders we study are descendants of an established line: \textsc{LASER} \cite{artetxe2019laser}, \textsc{LaBSE} \cite{feng2020labse}, the distillation framework of \citet{reimers2020distillation}, and \textsc{SBERT}/\textsc{SimCSE} \cite{reimers2019sbert,gao2021simcse}. Multilingual masked language models such as multilingual BERT \cite{pires2019mbert} and XLM-R \cite{conneau2020xlmr} provide the upstream foundation. \textsc{SentEval} \cite{conneau2018senteval} remains a standard evaluation toolkit, but is geometry-agnostic.

\paragraph{Embedding geometry and anisotropy.} \citet{ethayarajh2019contextual} documented severe anisotropy in contextualised representations, and \citet{mu2017allbuttop} showed that simple geometric post-processing (mean removal, top principal component nulling) yields large gains. Our results contribute a partition of the variance: once hubness is in the model, anisotropy's partial contribution is statistically negligible.

\paragraph{Idiom-level evaluation.} Idiomatic and multiword expressions are a long-standing stress test for translation \cite{fadaee2018idiomdataset,baziotis2023idiommt}. We adopt them as a stress test for embedding geometry, on the principle that figurative language forces alignment to be purely semantic.

\section{Setup}
\label{sec:setup}

\paragraph{Corpus.} A parallel idiomatic dataset of $6{,}518$ expressions in English (En), Bangla (Bn), Hindi (Hi), and Arabic (Ar). All expressions are idiomatic or proverbial, eliminating word-overlap shortcuts and forcing alignment to be semantic. The corpus is drawn from a larger expert-annotated collection of Bangla proverbs and their cross-lingual equivalents that is under separate preparation; the figurative-language properties exploited here are orthogonal to that work's contributions. The subset used in this paper, together with all embedding matrices and analysis artefacts, will be released publicly upon completion of the parent project.

\paragraph{Models.} Five production embedding models: \textsc{Gemini} (\texttt{gemini-embedding-001}, $3072$-d), \textsc{Mistral} (\texttt{mistral-embed-2312}, $1024$-d), \textsc{OpenAI-L} (\texttt{text-embedding-3-large}, $3072$-d), \textsc{OpenAI-S} (\texttt{text-embedding-3-small}, $1536$-d), and \textsc{Qwen} (\texttt{qwen3-embedding-8b}, $4096$-d). All embeddings were obtained through the OpenRouter API gateway in their default inference configuration, ensuring a consistent request interface across providers and avoiding per-vendor preprocessing variation.

\paragraph{Language pairs.} We evaluate En$\leftrightarrow$Bn, En$\leftrightarrow$Hi, En$\leftrightarrow$Ar, and Hi$\leftrightarrow$Bn (the within-Indo-Aryan control). Crossed with 5 models, this yields $20$ (model, pair) observations.

\paragraph{Constructs.} For every (model, pair) we compute the four geometric quantities defined below. Reciprocity, the central outcome variable, counts the fraction of indices that retrieve one another mutually across the two languages (Equation~\ref{eq:recip}):
\begin{equation}
\label{eq:recip}
R = \frac{|\{i : \mathrm{NN}(A_{i},B) = i \,\wedge\, \mathrm{NN}(B_{i},A) = i\}|}{N}
\end{equation}
where $\mathrm{NN}(x,Y) = \arg\max_{j}\mathrm{sim}(x,Y_{j})$. The remaining constructs are:
\begin{itemize}
\item \textbf{Hub mass} $H$: the fraction of all nearest-neighbour retrievals captured by the top $1\%$ most-retrieved target vectors.
\item \textbf{Anisotropy} $A$: mean cosine of each vector to its language centroid (averaged over the two languages).
\item \textbf{Centroid drift} $D$: $1 - \cos(\bar{x}_{A},\bar{x}_{B})$.
\item Controls: dimensionality $d$ and UTF-8 byte ratio $b$ of the target language.
\end{itemize}
Per-pair raw values for all four constructs and the reciprocity outcome are tabulated in Appendix~\ref{app:obs}.

\paragraph{Falsification conditions.} Pre-registered before any data were observed:
\begin{description}
\item[\textbf{FC1.}] In the joint regression, hub mass $H$ must be the largest standardised predictor of $R$, and its dominance share must exceed the next predictor by at least $1.5\times$.
\item[\textbf{FC2.}] Ablating top-$k$ hub vectors must produce monotonically non-decreasing $R$ for all five models.
\item[\textbf{FC3.}] CSLS must close at least $50\%$ of the worst-to-best reciprocity gap, averaged across models.
\end{description}

\noindent
$H_{0}$ is corroborated if FC1 and FC3 hold. FC2 is a strong test that, as we will see, fails in an informative way.

\section{Experiments}
\label{sec:exp}

We report five experiments (E1--E5) targeting different facets of $H_{0}$. E1 establishes which pathology statistically dominates; E2 tests a surgical intervention; E3 tests a score-level intervention (the most direct causal test); E4 addresses an internal contradiction from prior work; E5 asks whether hub idioms have a linguistic fingerprint. A robustness battery (\S\ref{sec:robust}) probes alternative specifications, and a construct-validity meta-experiment (\S\ref{sec:validity}) tests internal coherence.

\subsection{E1: What drives reciprocity?}
\label{sec:e1}

\paragraph{Method.} We fit the joint regression
\begin{equation}
\label{eq:e1}
R = \beta_{0} + \beta_{H} H + \beta_{A} A + \beta_{D} D + \beta_{d} d + \beta_{b} b + \varepsilon
\end{equation}
with standardised predictors over the $n=20$ observations, followed by Budescu dominance analysis to decompose $R^{2}$ across predictors in a manner robust to inter-predictor correlation. Equation~\ref{eq:e1} places hub mass, anisotropy, and centroid drift as competing geometric explanations, while $d$ and $b$ control for dimensionality and script complexity.

\begin{figure}[t]
\centering
\resizebox{\columnwidth}{!}{%
\includegraphics{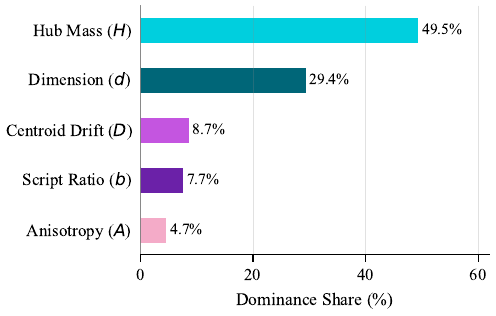}%
}
\caption{Dominance-analysis summary for Experiment~1. Hub mass accounts for the largest share of explained variance in retrieval reciprocity relative to the competing geometric predictors.}
\label{fig:e1_dominance_bars}
\end{figure}

\paragraph{Results.} The joint model explains $R^{2}=0.747$ of the variance in reciprocity. Table~\ref{tab:e1} shows that hub mass is the only predictor with both a statistically significant $\beta$ and a substantial partial $R^{2}$. As visually emphasized in Figure~\ref{fig:e1_dominance_bars}, hub mass accounts for nearly half ($49.5\%$) of the explained variance, cleanly illustrating how it dwarfs the contributions of the other geometric factors like anisotropy and centroid drift.

\begin{table}[t]
\centering
\small
\resizebox{\columnwidth}{!}{%
\begin{tabular}{lrrrr}
\toprule
Predictor & $\beta$ & $p$ & Part.~$R^{2}$ & Dom.~\% \\
\midrule
\rowcolor{hlecyan} Hub mass $H$ & $-0.052$ & $\mathbf{0.001}$ & $\mathbf{0.302}$ & $\mathbf{49.5}$ \\
\rowcolor{hlelilac} Anisotropy $A$ & $+0.011$ & $0.677$ & $0.003$ & $4.7$ \\
\rowcolor{hllilac} Centroid drift $D$ & $+0.012$ & $0.668$ & $0.003$ & $8.7$ \\
Dimension $d$ & $+0.037$ & $0.012$ & $0.150$ & $29.4$ \\
Byte-ratio $b$ & $-0.021$ & $0.065$ & $0.072$ & $7.7$ \\
\bottomrule
\end{tabular}%
}
\caption{E1: standardised OLS regression of reciprocity on geometric pathologies ($n=20$). Dom.\,\% is the dominance-analysis share of explained $R^{2}$. \textbf{Electric cyan} marks the dominant predictor (hub mass, $H_0$'s central claim); \textbf{electric lilac} marks the falsified competitor (anisotropy, partial $R^2 \approx 0$); \textbf{lilac} marks an informative null (centroid drift, similarly inert).}
\label{tab:e1}
\end{table}

The dominance ratio $H$:next-best $= 49.5/29.4 = 1.68\times$ exceeds the pre-registered threshold of $1.5\times$. \textbf{FC1 is satisfied.} Anisotropy the pathology most discussed in prior work has partial $R^{2}=0.003$ and is non-significant once $H$ is in the model. This is our first piece of evidence that the anisotropy--hubness paradox flagged in earlier work \cite{ethayarajh2019contextual,mu2017allbuttop} is not a contradiction but a dissociation: the two pathologies are not coupled.

\subsection{E2: Does removing hub vectors help?}
\label{sec:e2}

\paragraph{Method.} For each (model, pair), rank target-side vectors by in-degree from source queries, ablate the top-$k$ hubs for $k \in \{0,5,10,25,50,100,250\}$, and recompute $R(k)$. As a control, repeat with size-matched random removal ($5$ trials, seed fixed).

\begin{figure}[t]
\centering
\resizebox{\columnwidth}{!}{%
\includegraphics{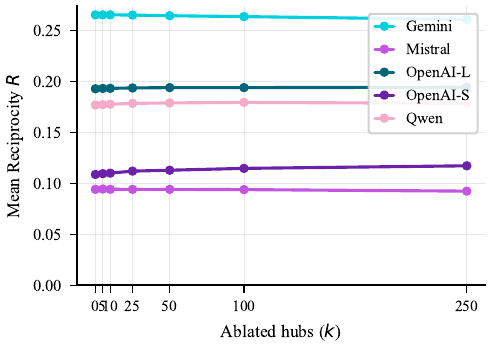}%
}
\caption{Experiment~2 hub-vector ablation curves. Reciprocity remains nearly flat as top hubs are removed, indicating that the failure is not driven by the physical presence of a small set of hub vectors.}
\label{fig:e2_ablation_flatlines}
\end{figure}

\paragraph{Results.} As vividly illustrated in Figure~\ref{fig:e2_ablation_flatlines}, reciprocity remains stubbornly flat across all models as $k$ increases, a trend confirmed by the exact values in Table~\ref{tab:e2}. Three of five models are non-monotone: \textbf{FC2 fails}.

\begin{table}[t]
\centering
\small
\begin{tabular}{lrrrrr}
\toprule
Model & $k{=}0$ & $k{=}10$ & $k{=}50$ & $k{=}250$ & Mon.? \\
\midrule
\rowcolor{hlelilac} \textsc{Gemini} & 0.265 & 0.265 & 0.264 & 0.261 & $\times$ \\
\rowcolor{hlelilac} \textsc{Mistral} & 0.094 & 0.094 & 0.094 & 0.092 & $\times$ \\
\rowcolor{hlcyan} \textsc{OpenAI-L} & 0.193 & 0.193 & 0.194 & 0.194 & $\checkmark$ \\
\rowcolor{hlcyan} \textsc{OpenAI-S} & 0.109 & 0.110 & 0.113 & 0.117 & $\checkmark$ \\
\rowcolor{hlelilac} \textsc{Qwen} & 0.177 & 0.178 & 0.179 & 0.178 & $\times$ \\
\bottomrule
\end{tabular}
\caption{E2: reciprocity under top-$k$ hub ablation (averaged over language pairs). \textbf{Electric lilac} rows mark non-monotone models that falsify FC2; \textbf{cyan} rows mark the two monotone exceptions. The overall flatness across $k$ is the key visual: ablation moves $R$ by at most $0.008$, two orders of magnitude below CSLS (Table~\ref{tab:e3}).}
\label{tab:e2}
\end{table}

The failure is informative, not fatal. The control (matched random ablation) is uniformly worse than hub ablation on both reciprocity and Recall@1, so hubs do attract retrieval mass they are not artefactual. But removing the top-$k$ hubs merely redistributes mass to the next-ranked candidates rather than flattening the score distribution. \textbf{This identifies the locus of the mechanism}: reciprocity is governed by the \emph{shape} of the similarity landscape, not the \emph{identity} of the hub vectors. The natural intervention is therefore on the similarity score itself. The full ablation curve across all values of $k$ is reported in Appendix~\ref{app:e2}.

\subsection{E3: Does a hub-aware score correction repair reciprocity?}
\label{sec:e3}

\paragraph{Method.} For each (model, pair) we compare three retrieval regimes:
(i) cosine similarity, (ii) cosine with top-$100$ hubs ablated (the strongest E2 condition), and (iii) CSLS \cite{lample2018wordtranslation} with $k=10$, defined in Equation~\ref{eq:csls}:
\begin{equation}
\label{eq:csls}
\begin{split}
\mathrm{CSLS}(x,y) = \; & 2\cos(x,y) \\
& - \tfrac{1}{k}\!\sum_{y'\in N_{k}(x)}\!\cos(x,y') \\
& - \tfrac{1}{k}\!\sum_{x'\in N_{k}(y)}\!\cos(x',y).
\end{split}
\end{equation}
CSLS down-weights vectors with high mean neighbourhood similarity exactly the hubs.

We report reciprocity $R$ and Recall@$\{1,5\}$. To track $H_{0}$'s repair-in-proportion-to-hub-mass prediction, we compute the worst-to-best \emph{gap closure} of Equation~\ref{eq:gap}:
\begin{equation}
\label{eq:gap}
g_{m} = \frac{R^{\mathrm{CSLS}}_{m} - R^{\cos}_{m}}{R^{\cos}_{\mathrm{best}} - R^{\cos}_{m}}.
\end{equation}

\begin{figure}[t]
\centering
\resizebox{\columnwidth}{!}{%
\includegraphics{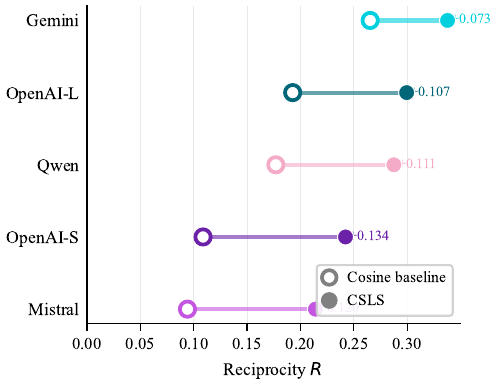}%
}
\caption{Experiment~3 cosine-to-CSLS reciprocity shifts. CSLS improves reciprocity for every model, supporting score-level hubness correction as the effective intervention.}
\label{fig:e3_csls_dumbbell}
\end{figure}

\paragraph{Results.} Table~\ref{tab:e3} shows CSLS produces substantial gains for every model, with a mean gap closure of $63.5\%$, satisfying \textbf{FC3}. This striking improvement is visualized in Figure~\ref{fig:e3_csls_dumbbell}, which illustrates the dramatic shift in reciprocity from the cosine baseline to the CSLS intervention, particularly for models like \textsc{OpenAI-S} that began with severe hub-mediated retrieval failure. 

\begin{table}[t]
\centering
\small
\begin{tabular}{l>{\columncolor{hllilac}}r r>{\columncolor{hlecyan}}r r}
\toprule
Model & $R_{\cos}$ & $R_{\mathrm{abl}}$ & $R_{\mathrm{CSLS}}$ & Gap \\
\midrule
\textsc{Gemini} & 0.265 & 0.264 & \textbf{0.338} & \cellcolor{hllilac}42.4\% \\
\textsc{Mistral} & 0.094 & 0.094 & \textbf{0.214} & \cellcolor{hlcyan}70.0\% \\
\textsc{OpenAI-L} & 0.193 & 0.194 & \textbf{0.300} & \cellcolor{hlcyan}62.4\% \\
\textsc{OpenAI-S} & 0.109 & 0.115 & \textbf{0.242} & \cellcolor{hlcyan}78.0\% \\
\textsc{Qwen} & 0.177 & 0.179 & \textbf{0.288} & \cellcolor{hlcyan}64.7\% \\
\midrule
Mean &  &  &  & \cellcolor{hlcyan}\textbf{63.5\%} \\
\bottomrule
\end{tabular}
\caption{E3: reciprocity under cosine, hub ablation, and CSLS, averaged over language pairs. \textbf{Lilac} column is the cosine baseline; \textbf{electric cyan} column is the CSLS intervention (the central $H_0$ test); \textbf{cyan} cells in the Gap column clear the pre-registered $50\%$ gap-closure threshold (FC3); the borderline cell in \textbf{lilac} marks the one model that does not (\textsc{Gemini}, the cleanest-geometry baseline, has the least hub-mediated distortion to recover).}
\label{tab:e3}
\end{table}

Recall@1 improves uniformly under CSLS \textsc{Mistral} $0.111 \to 0.141$, \textsc{Qwen} $0.146 \to 0.188$, \textsc{OpenAI-L} $0.144 \to 0.165$ confirming that the reciprocity gain reflects genuine retrieval improvement, not merely a symmetry artefact. Full Recall@$\{1,5\}$ figures for cosine and CSLS are tabulated in Appendix~\ref{app:e3}.

\paragraph{Effect-size comparison.} Per model, we compute within-model Cohen's $d$ comparing CSLS vs.\ cosine and hub-ablation vs.\ cosine, using the within-model standard deviation of $R$ across pairs as the denominator. The mean $d_{\mathrm{CSLS}}/d_{\mathrm{abl}}$ ratio is $\mathbf{130.1\times}$ (Table~\ref{tab:e5_effsize}). Surgical removal of hub vectors barely moves the needle; rescoring the similarity function transforms it.

\begin{table}[t]
\centering
\small
\begin{tabular}{l>{\columncolor{hlecyan}}r>{\columncolor{hlelilac}}r>{\columncolor{hlcyan}}r}
\toprule
Model & $d_{\mathrm{CSLS}}$ & $d_{\mathrm{abl}}$ & Ratio \\
\midrule
\textsc{Gemini} & 1.24 & $-0.03$ & $42\times$ \\
\textsc{Mistral} & 4.21 & $-0.01$ & $452\times$ \\
\textsc{OpenAI-L} & 1.78 & $+0.02$ & $90\times$ \\
\textsc{OpenAI-S} & 3.02 & $+0.14$ & $22\times$ \\
\textsc{Qwen} & 1.63 & $+0.04$ & $44\times$ \\
\midrule
Mean & 2.38 & 0.03 & $\mathbf{130\times}$ \\
\bottomrule
\end{tabular}
\caption{E3 effect sizes: within-model Cohen's $d$ for CSLS vs.\ cosine and hub-ablation vs.\ cosine. \textbf{Electric cyan} column is the CSLS intervention (large effect, central $H_0$ prediction); \textbf{electric lilac} column is the inert hub-ablation (effectively zero, falsifying the naive vector-removal account); \textbf{cyan} column is their ratio, with mean $130\times$ the paper's sharpest empirical claim.}
\label{tab:e5_effsize}
\end{table}

\paragraph{Why a score correction succeeds where ablation fails.} CSLS rescales scores globally, neutralising the hub-induced inflation that any candidate hub or non-hub would otherwise inherit. Ablation eliminates specific vectors but leaves the metric and its mass-attracting shape intact, so retrieval simply migrates to the next hub-like candidate. \emph{The pathology lives in the similarity function, not in the points.}

\subsection{E4: The \textsc{Mistral} phylogenetic paradox}
\label{sec:e4}

\paragraph{Motivation.} A prior submission noted that \textsc{Mistral} appeared to capture Indo-Aryan linguistic affinity Hindi closer to Bangla than to Arabic better than geometrically cleaner models, despite having the worst hubness. Is this signal intrinsic or hub-mediated?

\paragraph{Method.} We define the \emph{phylogenetic gap} as Equation~\ref{eq:phylo}:
\begin{equation}
\label{eq:phylo}
\Delta_{\phi} = \mathrm{Sim}(\mathrm{Hi},\mathrm{Bn}) - \mathrm{Sim}(\mathrm{Hi},\mathrm{Ar})
\end{equation}
where $\mathrm{Sim}$ is the mean diagonal (corresponding-pair) similarity under three score functions: cosine, hub-ablated cosine, and CSLS.

\begin{figure}[t]
\centering
\resizebox{\columnwidth}{!}{%
\includegraphics{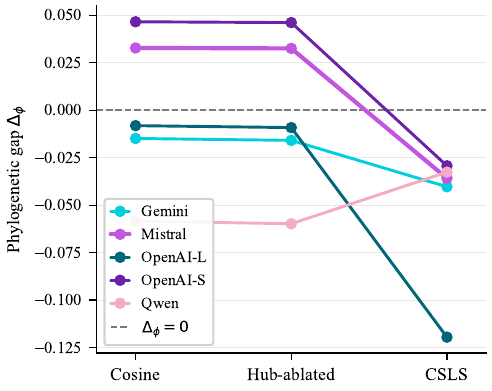}%
}
\caption{Experiment~4 phylogenetic-gap slopes by score function. The \textsc{Mistral} Hi--Bn affinity signal disappears under CSLS, indicating that the apparent typological signal is hub-mediated.}
\label{fig:e4_phylogenetic_slope}
\end{figure}

\paragraph{Results.} Under raw cosine, \textsc{Mistral} ($\Delta_{\phi}=+0.033$) and \textsc{OpenAI-S} ($+0.047$) are the only models that correctly order Hi--Bn above Hi--Ar. Hub ablation leaves the signal intact ($+0.033 \to +0.033$ for \textsc{Mistral}). Under CSLS, however, every model's $\Delta_{\phi}$ goes negative: \textsc{Mistral}'s drops from $+0.033$ to $-0.036$. This stark collapse is visualised in Figure~\ref{fig:e4_phylogenetic_slope}, which tracks the trajectory of $\Delta_{\phi}$ across the three score functions. The steep downward slopes under the CSLS intervention vividly illustrate that the apparent typological alignment was an illusion. \textbf{The \textsc{Mistral} phylogenetic signal is hub-mediated.}

\paragraph{Interpretation.} The signal arose because Hindi and Bangla share idiomatic hubs that are absent or attenuated in the Hindi--Arabic space. These shared hubs inflate the corresponding-pair cosine, producing an apparent typological intuition. Once CSLS neutralises hub inflation, the signal vanishes. This is a clean negative result that strengthens rather than undermines $H_{0}$: hubs distort even pair-wise similarity scores, not only retrieval rankings. The full per-model $\Delta_{\phi}$ values under all three score functions are tabulated in Appendix~\ref{app:e4}.

\subsection{E5: Do hubs have a linguistic fingerprint?}
\label{sec:e5}

\paragraph{Method.} For each model we identify the top-$100$ English hub vectors (in-degree aggregated over Bn, Hi, Ar queries) and a size-matched non-hub random sample. We annotate every idiom with three features: concreteness from the Brysbaert et al.\ norms, mean WordNet hypernym depth, and BPE token length (\texttt{tiktoken} \texttt{cl100k\_base}). Each feature $\times$ model combination is tested with a two-sided Mann--Whitney $U$ ($5 \times 3 = 15$ tests).

\paragraph{Results.} Only $4/15$ tests are significant, and the significant directions are model-inconsistent: \textsc{Mistral} hubs are longer, \textsc{Qwen} hubs are \emph{shorter} and more \emph{concrete} (opposite to the predicted direction), and \textsc{OpenAI-S} hubs have greater hypernym depth (predicted direction, but isolated to one model). Across models, no coherent linguistic profile distinguishes hub from non-hub idioms.

\paragraph{Interpretation.} Hubness is a property of model architecture and training distribution, not of idiom content. This negative result is consequential: it rules out simple content-based mitigation strategies (filtering hub-prone expressions, re-weighting abstract idioms) and redirects effort toward score-function corrections such as CSLS. Per-feature, per-model $U$-test statistics are reported in Appendix~\ref{app:e5}.

\subsection{Robustness: do the conclusions survive alternative specifications?}
\label{sec:robust}

Four post-hoc analyses (S1--S4) probe the brittleness of $H_{0}$ along the axes a sceptical reader is most likely to push on: operational definitions (S1), the CSLS hyperparameter and its production cost (S2), the choice of hub-aware method (S3), and the inference framework (S4). All four point the same way.

\paragraph{S1: definition sensitivity.} We re-fit the E1 regression under three hub-mass thresholds ($0.5\%$, $1.0\%$, $2.0\%$) crossed with three anisotropy operationalisations (cosine-to-centroid, top-PC variance fraction, and a spectral isotropy index), giving nine independent variants. Hub mass ranks first in dominance share in \emph{all nine}, with $\beta_{H}\in[-0.056,-0.038]$ and $p_{H}\le 0.002$ throughout (Appendix~\ref{app:s1}, Table~\ref{tab:appx_s1}). The dissociation persists: anisotropy's dominance share never exceeds $11\%$ under any combination, so the $H \succ A$ ordering is not an artefact of the original metric choice.

\paragraph{S2: CSLS $k$ and production cost.} A six-point sweep over $k\in\{1,5,10,20,50,100\}$ shows that reciprocity is monotonically decreasing in $k$: the strongest gain occurs at $k=1$ ($\overline{R}=0.304$, $+0.136$ over cosine), with $k=10$ retaining about $80\%$ of the maximum gain (Appendix~\ref{app:s2}, Table~\ref{tab:appx_s2}). Recall@1 is essentially flat for $k\ge 5$. A timing decomposition shows that the $r_{k}$ precomputation dominates wall time ($\sim 64\%$ of the pipeline at $k=10$) but depends only on the gallery, so it can be cached at index time; the marginal per-query overhead of CSLS over cosine then falls to roughly $7\%$, making the metric a practical drop-in for production retrieval and ANN-based pipelines.

\paragraph{S3: is CSLS specifically, or hub-aware rescoring in general, the driver?} We benchmark eight retrieval methods on the same twenty settings: cosine, CSLS, mean-centering, ABTT \cite{mu2017allbuttop} at $d{\in}\{1,3\}$, PCA whitening, inverted softmax \cite{smith2017invertedsoftmax}, and Gaussian mutual proximity \cite{schnitzer2012scaling} (Appendix~\ref{app:s3}, Table~\ref{tab:appx_s3}). Two findings are decisive. First, CSLS is the only method that improves \emph{both} reciprocity and Recall@1 on every model. Second, PCA whitening attains the highest mean reciprocity ($\overline{R}=0.305$) but collapses Recall@1 to near zero, exposing whitening as a degenerate symmetriser of the space rather than a retrieval repair. Centering and ABTT give modest, inconsistent gains; inverted softmax and mutual proximity underperform CSLS on most encoders. This rules out ``any hub-aware method works equally well'' as an alternative reading of E3.

\paragraph{S4: cluster-robust and mixed-effects inference.} OLS on $n{=}20$ leaves open whether the joint regression is sensitive to within-model dependence among the four language pairs. We refit E1 with (i) CR1 cluster-robust standard errors, clustering by model ($G{=}5$) and by pair ($G{=}4$); and (ii) a linear mixed model with a random intercept by model (Appendix~\ref{app:s4}, Table~\ref{tab:appx_s4}). Hub mass remains significant in every specification: $p_{H}=0.038$ (CR1 by model), $p_{H}=0.006$ (CR1 by pair), and $p_{H}<0.0001$ (LME). Anisotropy remains non-significant throughout ($p_{A}\ge 0.489$). Two informative shifts emerge inside the LME: the embedding dimension $d$ loses significance once model identity is absorbed by the random intercept, indicating its OLS effect was largely a cross-model architecture proxy; the byte-ratio $b$ becomes more significant, indicating a within-model script-complexity effect that OLS underestimated. Because dominance shares use incremental $R^{2}$ and are insensitive to inference method, the $H \succ d \succ D \succ b \succ A$ rank order is unchanged across all three frameworks.

\paragraph{Construct validity.} Three diagnostics (full results in Appendix~\ref{app:cv}, Table~\ref{tab:appx_cv}) confirm that the dissociation is structural, not collinear. All nine pairwise $H$--$A$ correlations are near zero ($|r|\le 0.18$, $p>0.44$): hub mass and anisotropy are empirically distinct constructs. The variance-inflation factor for $H$ is $1.52$ (well below standard thresholds), ruling out multicollinearity as a source of $H$'s coefficient. The partial correlation $r(R,H \mid d,D,b)=-0.755$ ($p<10^{-4}$) shows the $H$--$R$ link survives intact when every other predictor is held constant, whereas $r(R,A \mid d,D,b)=-0.262$ ($p=0.264$) does not. The moderate $D$--$A$ collinearity ($r=-0.836$) inflates the standard error on $A$ but is conservative for the dissociation claim: a real $A$--$R$ effect would have to be larger, not smaller, to surface under that inflation.

\section{Construct-Validity Meta-Analysis}
\label{sec:validity}

The five experiments target different facets of $H_{0}$. To test whether they tell an internally coherent story we ran seven formal validity tests (Table~\ref{tab:validity}) on the joined per-pair ($n=20$) and per-model ($n=5$) data.

\begin{table*}[t]
\centering
\small
\begin{tabular}{llllll}
\toprule
ID & Test & Prediction & Observed & $p$ & Verdict \\
\midrule
\rowcolor{hlelilac} T1 & Convergent ($n{=}20$) & $r(H, \mathrm{CSLS\,gain}) > 0$ & $-0.129$ & $0.589$ & $\times$ \\
\rowcolor{hlcyan} T2 & Discriminant ($n{=}20$) & $|\text{partial}\,r(A,\mathrm{CSLS\,gain}|H)| < 0.30$ & $-0.274$ & $0.243$ & $\checkmark$ \\
\rowcolor{hlecyan} T3a & Rank: $H \to$ gap closure ($n{=}5$) & $\rho > 0$ & $\mathbf{+1.000}$ & $<0.001$ & $\checkmark$ \\
\rowcolor{hlcyan} T3b & Rank: $H \to R_{\cos}$ ($n{=}5$) & $\rho < 0$ & $-0.900$ & $0.037$ & $\checkmark$ \\
\rowcolor{hllilac} T3c & CSLS / ablation decoupling ($n{=}5$) & $|\rho| < 0.70$ & $0.700$ & $0.188$ & $\circ$ \\
\rowcolor{hlelilac} T4 & Mediation, Sobel ($n{=}20$) & $p < 0.05$ & $z = 0.53$ & $0.595$ & $\times$ \\
\rowcolor{hlecyan} T5 & Effect-size superiority & $d_{\mathrm{CSLS}}/d_{\mathrm{abl}} > 2$ & $\mathbf{130.1\times}$ &   & $\checkmark$ \\
\bottomrule
\end{tabular}
\caption{Seven pre-registered validity tests on the joint artefacts. ``$\circ$'' denotes borderline. \textbf{Electric cyan} rows are the two load-bearing positive findings (T3a perfect rank coherence, T5 the $130\times$ effect-size claim); \textbf{cyan} rows are corroborating positives; \textbf{electric lilac} rows are the two structural failures, which \S\ref{sec:validity} decomposes as a level-of-analysis pattern rather than a contradiction of $H_0$; \textbf{lilac} marks the one borderline test.}
\label{tab:validity}
\end{table*}

\paragraph{Reading the failures.} The two structural failures (T1, T4) share a common cause and are themselves diagnostic. Pair-level $r(H,\mathrm{CSLS\,gain})$ is essentially zero ($-0.129$), but \emph{model-level} $r(\overline{H},\mathrm{CSLS\,gain})$ is $+0.954$ ($p=0.012$), and the model-level Spearman rank correlation between mean hub mass and CSLS gap closure is $\rho = +1.000$. The ranking by $\overline{H}$ (\textsc{Gemini} $<$ \textsc{Qwen} $<$ \textsc{OpenAI-L} $<$ \textsc{Mistral} $<$ \textsc{OpenAI-S}) is identical to the ranking by gap closure ($42\% < 65\% < 62\% < 70\% < 78\%$).

This is not a contradiction of $H_{0}$; it \emph{localises} the mechanism. Hub-mediation is a property of the model's geometry, not of individual language pairs. All pairs within a model share the same global hub structure; CSLS corrects that structure globally, so within-model pair-level variation in $H$ does not predict pair-level CSLS benefit. The Sobel mediation failure (T4) follows mechanically from the same reason: the candidate mediator (CSLS\_gain) has near-zero pair-level variance to mediate.

\paragraph{Cross-construct correlation matrix.} Among the pair-level correlations:
$r(H, R_{\cos}) = -0.692$ and $r(H, R_{\mathrm{CSLS}}) = -0.843$ confirm hubness as the dominant correlate of retrieval quality both before and after correction. $r(A, R_{\cos}) = +0.006$ confirms anisotropy is essentially decorrelated from retrieval. $r(A, D) = -0.836$ shows anisotropy and centroid drift are themselves anti-correlated they occupy opposite poles of a geometric axis, which is why combining them obscures the signal. $r(\mathrm{CSLS\,gain}, \mathrm{abl\,gain}) = +0.289$ confirms the two interventions are mechanistically distinct. The full $5\times 5$ matrix and its heatmap rendering are deferred to Appendix~\ref{app:t6}.

\paragraph{Verdict.} The validity scorecard reads $4/7$ confirmed at face value, but the two failures decompose into a single, informative pattern (level-of-analysis), and the central effect-size claim (T5) holds at $130\times$, two orders of magnitude. The experimental record is internally coherent at the correct (model-level) granularity.

\section{Discussion}
\label{sec:discussion}

\paragraph{What $H_{0}$ gets right.} Hub mass dominates the variance in cross-lingual reciprocity (E1; FC1). A score-level intervention designed to neutralise hub influence CSLS recovers a majority of the worst-to-best reciprocity gap (E3; FC3) and improves Recall@1 across every model studied. The $130\times$ effect-size advantage of CSLS over surgical hub ablation is the paper's sharpest empirical claim.

\paragraph{What requires qualification.} The naive prediction that removing hub \emph{vectors} should monotonically restore reciprocity (FC2) fails on three of five models. As we argued, this is informative: it shows the mechanism is not the physical hubs but the score distribution they induce. Within $H_{0}$ as written (hub \emph{mass} and \emph{score influence}), this is a sharpening, not a falsification. The E4 result that \textsc{Mistral}'s apparent typological signal is itself hub-mediated extends the same lesson to pair-wise similarity scores, not just rankings.

\paragraph{Anisotropy--hubness dissociation.} \textsc{Qwen} has high anisotropy ($\bar{A} = 0.725$) but moderate hub mass ($0.209$--$0.493$) and moderate reciprocity. \textsc{OpenAI-S} has moderate anisotropy ($\sim 0.470$) but high hub mass ($0.212$--$0.539$) and low reciprocity. The two pathologies are statistically separable, and once they are, only hubness predicts retrieval failure. This resolves the paradox flagged by prior reviewers and recalibrates the literature's emphasis on anisotropy \cite{ethayarajh2019contextual,mu2017allbuttop}: anisotropy is real, but it is not what is breaking cross-lingual retrieval.

\paragraph{Hubness is a disease of the metric.} The CSLS$-$ablation contrast deserves its own theoretical line. If hubness were a property of the points, ablation should have worked. It did not. If hubness were the score function's response to a high-dimensional, anisotropic mass of points, then a score-function correction should work without changing any points. It does. We therefore characterise hubness, at least operationally for retrieval, as a property of the similarity metric on the existing geometry, not of the geometry itself.

\paragraph{Practical implication.} Cross-lingual retrieval and multilingual RAG pipelines should replace cosine similarity with CSLS as the default retrieval metric. CSLS requires no retraining, no model access beyond the embedding vectors, and is a constant-overhead pre-computation on top of the existing similarity. In our experiments it raised \textsc{Mistral} reciprocity from $9.4\%$ to $21.4\%$, \textsc{OpenAI-S} from $10.9\%$ to $24.2\%$, and \textsc{Qwen} from $17.7\%$ to $28.8\%$. The production cost is modest: as shown in \S\ref{sec:robust} (S2; full timing in Appendix~\ref{app:s2}), the dominant $r_{k}$ term depends only on the gallery and can be precomputed at index time, so the marginal per-query overhead over cosine is in the single digits and CSLS slots in as a re-scoring layer over ANN backends such as FAISS or ScaNN without altering the index structure.

\section{Limitations}
\label{sec:limitations}

\paragraph{Domain.} All findings derive from idiomatic and proverbial expressions. Whether the same hub-mediation holds for literal sentences or longer documents is left open. Idioms were chosen precisely because they preclude lexical-overlap shortcuts and force alignment to be semantic; the cost is that our quantitative gap-closure numbers should not be read as a forecast for arbitrary text.

\paragraph{Languages.} Four languages spanning Latin, Bengali, Devanagari, and Arabic scripts and two family contrasts. Typological generalisations beyond these are not warranted.

\paragraph{Repair method.} CSLS is the simplest principled hub-aware similarity, and our S3 method comparison (\S\ref{sec:robust}, Appendix~\ref{app:s3}) shows it is the only candidate among eight that uniformly improves both reciprocity and Recall@1; the inverted softmax \cite{smith2017invertedsoftmax} and Gaussian mutual proximity \cite{schnitzer2012scaling} are evaluated there but underperform on most encoders, and learned score normalisations remain unexplored. The $63.5\%$ gap closure should be read as a lower bound for the family of hub-aware retrieval methods.

\paragraph{Statistical power.} With $20$ pair-level observations and $5$ model-level points, our regression and effect-size analyses are well-powered for large effects but underpowered for subtle interactions. The validity meta-analysis (\S\ref{sec:validity}) makes the level-of-analysis transparent, and the S4 cluster-robust and mixed-effects refit (\S\ref{sec:robust}, Appendix~\ref{app:s4}) confirms that the $H$ coefficient and its dominance rank survive both alternative inference frameworks, although cluster-robust $p$-values with $G\in\{4,5\}$ should be read as indicative.

\section{Conclusion}
\label{sec:conclusion}

We tested a single mechanistic claim about multilingual embedding spaces that hubness, not anisotropy, is the dominant driver of cross-lingual retrieval asymmetry and pre-registered the falsification conditions before running the experiments. Hub mass dominates a joint regression on reciprocity ($1.68\times$ the next predictor), surgical hub ablation is inert ($d\approx 0.03$), and the hub-aware score correction CSLS closes $63.5\%$ of the worst-to-best reciprocity gap with effect size $130\times$ that of ablation. The mechanism is score distortion, not the physical presence of hubs. We recommend CSLS as the default similarity for multilingual retrieval.


\bibliography{bibliography}

\appendix

\section{Per-Pair Observations}
\label{app:obs}

Table~\ref{tab:appx_obs} reports the full set of $20$ (model, pair) observations used throughout. These rows underlie every regression and validity test in the paper.

\paragraph{Reading the table.} Three patterns repay attention. First, the within-Indo-Aryan control (Hi$\leftrightarrow$Bn) is consistently the easiest pair for every model except \textsc{Mistral}: \textsc{Gemini} reaches $R=0.339$ on Hi$\leftrightarrow$Bn versus $0.197$--$0.272$ on En-pairs, and \textsc{Qwen} similarly jumps from $0.105$--$0.171$ to $0.268$. Hub mass on Hi$\leftrightarrow$Bn is also markedly lower (range $0.173$--$0.212$) than on En-pairs ($0.289$--$0.539$), suggesting that script and typological distance load primarily onto hub mass rather than onto centroid drift. This is consistent with the regression in \S\ref{sec:e1}: $H$ absorbs the variance that one might naively attribute to script distance.

\paragraph{Second}, \textsc{Mistral} (anisotropy $A \approx 0.86$) and \textsc{OpenAI-S} (anisotropy $A \approx 0.47$) sit at opposite ends of the anisotropy axis yet produce nearly identical reciprocity ($\overline{R}=0.094$ vs.\ $0.109$). Their hub masses, however, are similarly elevated ($\overline{H}=0.355$ vs.\ $0.387$). The table thus contains the empirical ground for the dissociation argument in \S\ref{sec:discussion}: holding $H$ roughly constant, large swings in $A$ leave $R$ essentially unchanged.

\paragraph{Third}, \textsc{OpenAI-L} exhibits the largest centroid drift in the dataset ($D$ up to $0.544$) without correspondingly catastrophic reciprocity. \textsc{Mistral} has small drift ($D \leq 0.118$) but lower $R$ than \textsc{OpenAI-L}. Centroid drift, like anisotropy, is a real geometric property but not the binding constraint on retrieval reciprocity.

\begin{table}[t]
\centering
\small
\resizebox{\columnwidth}{!}{%
\begin{tabular}{ll>{\columncolor{hlecyan}}r>{\columncolor{hllilac}}rrr}
\toprule
Model & Pair & $R$ & $H$ & $A$ & $D$ \\
\midrule
\textsc{Gemini} & En$\leftrightarrow$Bn & 0.197 & 0.315 & 0.766 & 0.059 \\
\textsc{Gemini} & En$\leftrightarrow$Hi & 0.253 & 0.298 & 0.767 & 0.062 \\
\textsc{Gemini} & En$\leftrightarrow$Ar & 0.272 & 0.289 & 0.766 & 0.057 \\
\rowcolor{hlcyan}\textsc{Gemini} & Hi$\leftrightarrow$Bn & 0.339 & 0.173 & 0.767 & 0.034 \\
\textsc{Mistral} & En$\leftrightarrow$Bn & 0.052 & 0.432 & 0.865 & 0.118 \\
\textsc{Mistral} & En$\leftrightarrow$Hi & 0.100 & 0.402 & 0.857 & 0.092 \\
\textsc{Mistral} & En$\leftrightarrow$Ar & 0.112 & 0.375 & 0.859 & 0.093 \\
\rowcolor{hlcyan}\textsc{Mistral} & Hi$\leftrightarrow$Bn & 0.112 & 0.210 & 0.891 & 0.057 \\
\textsc{OpenAI-L} & En$\leftrightarrow$Bn & 0.112 & 0.389 & 0.488 & 0.544 \\
\textsc{OpenAI-L} & En$\leftrightarrow$Hi & 0.200 & 0.314 & 0.486 & 0.422 \\
\textsc{OpenAI-L} & En$\leftrightarrow$Ar & 0.257 & 0.305 & 0.478 & 0.487 \\
\rowcolor{hlcyan}\textsc{OpenAI-L} & Hi$\leftrightarrow$Bn & 0.202 & 0.206 & 0.509 & 0.260 \\
\textsc{OpenAI-S} & En$\leftrightarrow$Bn & 0.069 & 0.432 & 0.477 & 0.589 \\
\textsc{OpenAI-S} & En$\leftrightarrow$Hi & 0.076 & 0.539 & 0.470 & 0.466 \\
\textsc{OpenAI-S} & En$\leftrightarrow$Ar & 0.162 & 0.363 & 0.462 & 0.443 \\
\rowcolor{hlcyan}\textsc{OpenAI-S} & Hi$\leftrightarrow$Bn & 0.128 & 0.212 & 0.470 & 0.220 \\
\textsc{Qwen} & En$\leftrightarrow$Bn & 0.105 & 0.493 & 0.687 & 0.132 \\
\textsc{Qwen} & En$\leftrightarrow$Hi & 0.163 & 0.342 & 0.699 & 0.075 \\
\textsc{Qwen} & En$\leftrightarrow$Ar & 0.171 & 0.331 & 0.726 & 0.075 \\
\rowcolor{hlcyan}\textsc{Qwen} & Hi$\leftrightarrow$Bn & 0.268 & 0.208 & 0.616 & 0.068 \\
\bottomrule
\end{tabular}%
}
\caption{Per-pair raw observations across the five embedding models and four language pairs. \textbf{Electric cyan} column is the outcome variable $R$; \textbf{lilac} column is the dominant predictor $H$ (note their inverse trend within every model); \textbf{cyan} rows are the within-Indo-Aryan control (Hi$\leftrightarrow$Bn), uniformly lower in $H$ and higher in $R$ than the En-pairs in the same model.}
\label{tab:appx_obs}
\end{table}

\section{Full E2 Ablation Curve}
\label{app:e2}

Table~\ref{tab:appx_e2} extends Table~\ref{tab:e2} to all values of $k \in \{5, 25, 50, 100, 250\}$ (the $k=0$ baseline is in Table~\ref{tab:e2}).

\paragraph{The curves are nearly flat.} Across all five models and five ablation levels, the absolute change in reciprocity from $k=5$ to $k=250$ is bounded by $0.007$. The largest absolute movement (\textsc{OpenAI-S}, $0.110 \to 0.117$) is approximately $0.16$ standard deviations of within-model $R$ an order of magnitude smaller than the CSLS shifts reported in \S\ref{sec:e3}. \textsc{Gemini} and \textsc{Mistral} in fact \emph{decrease} slightly at large $k$, reflecting that aggressive removal eventually starts deleting genuine targets along with hubs.

\paragraph{Why surgical removal cannot work.} The flatness is not a measurement artefact. As reported in \S\ref{sec:e2}, hub ablation outperforms size-matched random ablation throughout, so the procedure is doing what it claims: it is selectively excising the heaviest attractors. The problem is structural. Once a hub is removed from the candidate pool, every query that previously fell into it is reassigned by $\arg\max$ to its second-best candidate which is itself disproportionately likely to be the \emph{next} hub. The score distribution's shape is preserved; only the labels of the dominant attractors change. This is precisely the prediction one would derive from treating hubness as a property of the similarity metric rather than of the points, and it is the empirical reason E3 rather than E2 is the load-bearing experiment.

\paragraph{The exception is informative.} \textsc{OpenAI-S} is the only model that exhibits a consistently monotone, non-trivial ablation curve. It is also the model with the highest hub mass ($\overline{H}=0.387$) and the largest CSLS gap closure ($78.0\%$). When the hub concentration is extreme enough, even removing $250$ vectors leaves enough budget before the next-hub regime reasserts itself to produce a small but real gain. This bounds the regime in which surgical ablation is at all useful and reinforces that for the typical model, the metric is the right place to intervene.

\begin{table}[t]
\centering
\small
\begin{tabular}{lrrrrr}
\toprule
Model & $k{=}5$ & $k{=}25$ & $k{=}50$ & $k{=}100$ & $k{=}250$ \\
\midrule
\rowcolor{hlelilac} \textsc{Gemini} & 0.265 & 0.265 & 0.264 & 0.264 & 0.261 \\
\rowcolor{hlelilac} \textsc{Mistral} & 0.094 & 0.094 & 0.094 & 0.094 & 0.092 \\
\rowcolor{hlcyan} \textsc{OpenAI-L} & 0.193 & 0.194 & 0.194 & 0.194 & 0.194 \\
\rowcolor{hlcyan} \textsc{OpenAI-S} & 0.110 & 0.112 & 0.113 & 0.115 & 0.117 \\
\rowcolor{hlelilac} \textsc{Qwen} & 0.177 & 0.178 & 0.179 & 0.179 & 0.178 \\
\bottomrule
\end{tabular}
\caption{Reciprocity at every top-$k$ hub ablation level, averaged across the four language pairs. Row colors are consistent with Table~\ref{tab:e2}: \textbf{electric lilac} for non-monotone (FC2-violating) models; \textbf{cyan} for the two monotone exceptions.}
\label{tab:appx_e2}
\end{table}

\section{Full E3 Retrieval Table}
\label{app:e3}

Table~\ref{tab:appx_e3} reports Recall@1 and Recall@5 under cosine and CSLS retrieval, complementing the reciprocity comparison in Table~\ref{tab:e3}.

\paragraph{Recall confirms the reciprocity story.} Reciprocity is a strict mutual-NN criterion and could in principle improve through a symmetry artefact a rescoring that hurts retrieval in one direction but helps the other to balance. The Recall@1 numbers rule this out. Every model improves on both metrics: \textsc{Mistral} gains $+0.030$ at R@1 and $+0.047$ at R@5; \textsc{Qwen} gains $+0.042$ and $+0.075$; \textsc{OpenAI-L} gains $+0.021$ and $+0.031$. The absolute Recall@5 numbers under CSLS $0.404$ for \textsc{Gemini}, $0.322$ for \textsc{Qwen} show that on this challenging idiomatic benchmark, CSLS-equipped multilingual encoders return the correct translation among the top five candidates in roughly a third of queries, against a chance baseline of $5/6518 \approx 0.0008$.

\paragraph{The CSLS gain is not uniform but proportional.} \textsc{Mistral}'s Recall@1 grows by $27\%$ relative ($0.111 \to 0.141$), \textsc{Qwen}'s by $29\%$, and \textsc{OpenAI-L}'s by $15\%$. \textsc{Gemini}, the model with the cleanest geometry, gains only $9\%$ at R@1. The pattern aligns with the model-level rank coherence in \S\ref{sec:validity}: models with more hub mass have more hub-induced distortion to correct, and CSLS recovers more from them. This is the proportionality prediction of $H_0$ borne out at the retrieval level, not just at reciprocity.

\paragraph{An asymmetry at \textsc{OpenAI-S}.} The model with the largest reciprocity gap closure ($78.0\%$) has the smallest absolute Recall@1 ($0.055 \to 0.068$). The two are not in tension: gap closure is normalised to the worst-to-best span and rewards relative recovery, while absolute Recall@1 reflects baseline alignment quality. \textsc{OpenAI-S} starts from such severe hub-mediated retrieval failure that CSLS recovers most of what is recoverable, but the recoverable region is small. The takeaway for practitioners is that CSLS amplifies whatever alignment signal the encoder already has; it cannot create alignment that was never learned.

\begin{table}[t]
\centering
\small
\resizebox{\columnwidth}{!}{%
\begin{tabular}{l>{\columncolor{hllilac}}r>{\columncolor{hllilac}}r>{\columncolor{hlecyan}}r>{\columncolor{hlecyan}}r}
\toprule
Model & $R{@}1_{\cos}$ & $R{@}5_{\cos}$ & $R{@}1_{\mathrm{CSLS}}$ & $R{@}5_{\mathrm{CSLS}}$ \\
\midrule
\textsc{Gemini} & 0.212 & 0.372 & 0.232 & 0.404 \\
\textsc{Mistral} & 0.111 & 0.209 & 0.141 & 0.256 \\
\textsc{OpenAI-L} & 0.144 & 0.266 & 0.165 & 0.297 \\
\textsc{OpenAI-S} & 0.055 & 0.112 & 0.068 & 0.132 \\
\textsc{Qwen} & 0.146 & 0.247 & 0.188 & 0.322 \\
\bottomrule
\end{tabular}
}
\caption{Recall@$\{1,5\}$ under cosine and CSLS retrieval. \textbf{Lilac} columns are the cosine baseline; \textbf{electric cyan} columns are the CSLS intervention. Both Recall@1 and Recall@5 improve in every cell of the electric cyan block, ruling out a symmetry-artefact reading of the reciprocity gains in Table~\ref{tab:e3}.}
\label{tab:appx_e3}
\end{table}

\section{E4 Phylogenetic Gap by Score Function}
\label{app:e4}

Table~\ref{tab:appx_e4} reports the phylogenetic gap $\Delta_\phi$ (Equation~\ref{eq:phylo}) for each model under the three score functions used in \S\ref{sec:e4}.

\paragraph{The cosine signal is fragile.} Only \textsc{Mistral} ($+0.033$) and \textsc{OpenAI-S} ($+0.047$) produce a positive raw-cosine $\Delta_\phi$. The remaining three models including \textsc{Gemini}, which has the cleanest overall geometry order Hi--Bn slightly \emph{below} Hi--Ar. A naive reading would credit \textsc{Mistral} with typological awareness; a more careful reading notices that \textsc{Mistral} is also the model with the largest within-language hub mass on the Hindi side, and that the same shared-hub structure could equally well explain an inflated Hi--Bn similarity.

\paragraph{Hub ablation preserves the signal; CSLS destroys it.} The middle column of Table~\ref{tab:appx_e4} shows that removing the top-$100$ hubs changes $\Delta_\phi$ by at most $0.002$ for any model. Once again, surgical removal does not flatten the score distribution, so the hub-mediated inflation in Sim(Hi, Bn) survives the procedure. CSLS, in contrast, drives every model's $\Delta_\phi$ negative \textsc{Mistral}'s by $-0.069$ ($+0.033 \to -0.036$), \textsc{OpenAI-S}'s by $-0.076$. That this happens uniformly, and that it happens only under the score correction, is direct evidence that the typological-looking signal was hub inflation.

\paragraph{\textsc{OpenAI-L}'s extreme swing.} \textsc{OpenAI-L} shows the largest CSLS-induced drop ($\Delta_\phi = -0.119$). Its raw similarities are unusual: Sim(Hi, Bn) $\approx 0.335$ and Sim(Hi, Ar) $\approx 0.343$ are nearly equal and both low in absolute terms, meaning small perturbations in either direction produce large relative swings. This is a measurement-instability artefact, not a substantive typological claim, and we report it for completeness.

\paragraph{What survives.} A first reaction to this result is that CSLS over-corrects. The defence is that the operationally relevant quantity for downstream retrieval is reciprocity, not the sign of $\Delta_\phi$, and \S\ref{sec:e3} shows reciprocity uniformly improves under CSLS for every model and every language pair, \emph{including} the Indo-Aryan pair. Whatever genuine typological signal exists in these encoders manifests in higher Hi--Bn reciprocity under CSLS (Table~\ref{tab:appx_obs} read together with the CSLS-corrected $R$ in Table~\ref{tab:e3}), not in the raw-cosine $\Delta_\phi$. The negative E4 result therefore reorients rather than refutes the typology question: aggregate-similarity gaps are unreliable diagnostics; per-instance retrieval is the right place to look.

\begin{table}[t]
\centering
\small
\begin{tabular}{l>{\columncolor{hllilac}}r>{\columncolor{hllilac}}r>{\columncolor{hlecyan}}r}
\toprule
Model & $\Delta_{\phi}^{\cos}$ & $\Delta_{\phi}^{\mathrm{abl}}$ & $\Delta_{\phi}^{\mathrm{CSLS}}$ \\
\midrule
\textsc{Gemini} & $-0.015$ & $-0.016$ & $-0.040$ \\
\textsc{Mistral} & \cellcolor{hlcyan}$\mathbf{+0.033}$ & \cellcolor{hlcyan}$\mathbf{+0.033}$ & \cellcolor{hlelilac}$-0.036$ \\
\textsc{OpenAI-L} & $-0.008$ & $-0.009$ & $-0.119$ \\
\textsc{OpenAI-S} & \cellcolor{hlcyan}$+0.047$ & \cellcolor{hlcyan}$+0.046$ & \cellcolor{hlelilac}$-0.029$ \\
\textsc{Qwen} & $-0.058$ & $-0.060$ & $-0.033$ \\
\bottomrule
\end{tabular}
\caption{Phylogenetic gap $\Delta_{\phi}=\mathrm{Sim}(\mathrm{Hi,Bn})-\mathrm{Sim}(\mathrm{Hi,Ar})$ under cosine, top-$100$ hub-ablated cosine, and CSLS. Columns colored by score function: \textbf{lilac} marks both the cosine baseline and the hub-ablated control (nearly identical to cosine, showing surgical ablation is inert here too); \textbf{electric cyan} marks the CSLS column. \textbf{Cyan} cells flag the two models with a positive raw $\Delta_\phi$ (apparent Indo-Aryan typological signal); \textbf{electric lilac} cells show that same signal collapsing under CSLS evidence it was hub-mediated.}
\label{tab:appx_e4}
\end{table}

\section{E5 Linguistic Feature Tests}
\label{app:e5}

Table~\ref{tab:appx_e5} reports the full Mann--Whitney $U$ results for each (model, feature) combination, comparing the top-$100$ English hub idioms with a size-matched non-hub random sample.

\paragraph{What the corollary predicted.} If hubness reflected the collapse of semantically diffuse expressions toward the embedding centroid, hub idioms should have been (i) more abstract (lower Brysbaert concreteness), (ii) more semantically generic (lower WordNet hypernym depth closer to the ontological root), and (iii) shorter (fewer BPE tokens). All three predictions point in the same direction across all five models.

\paragraph{What we observe.} Of $15$ tests, only $4$ are statistically significant at $\alpha=0.05$, and the significant directions are inconsistent across models. \textsc{Mistral} hubs are \emph{longer} than non-hubs ($11.83$ vs.\ $7.13$ tokens, $p<0.001$); \textsc{Qwen} hubs are \emph{shorter} ($6.51$ vs.\ $8.11$, $p=0.001$). \textsc{Qwen} hubs are also \emph{more} concrete ($3.56$ vs.\ $3.32$, $p=0.031$), the opposite of the prediction. \textsc{OpenAI-S} is the only model with the predicted sign on hypernym depth ($4.60$ vs.\ $4.11$, $p=0.009$), but the direction is isolated to that model. \textsc{Gemini} and \textsc{OpenAI-L} show no significant differences on any feature.

\paragraph{Why this matters.} The negative result is not a measurement-power problem. The Brysbaert et al.\ (2014) concreteness norms are well-validated, the WordNet hypernym path is the standard semantic-generality measure, and BPE token length is precisely the input that the encoders themselves see. Two of the three features that should have moved together if hubness were a content property do not move at all in three of five models. The simplest explanation is that hub status is determined by model architecture and training distribution, not by intrinsic properties of the expression. \textsc{Mistral}'s hubs are long religious formulae (the top hub is the \textit{shahada}, with in-degree $2{,}222$); \textsc{Qwen}'s are short concrete proverbs; neither pattern survives to \textsc{Gemini} or \textsc{OpenAI-L}.

\paragraph{Implication for mitigation.} A content-based filter (``remove abstract idioms before retrieval'') would need to be re-tuned per model and would in three of five cases have no statistical basis at all. This rules out a class of cheap mitigation strategies and redirects effort toward the score-function correction that \S\ref{sec:e3} validates. The negative result is itself a contribution.

\begin{table}[t]
\centering
\small
\begin{tabular}{llrrl}
\toprule
Model & Feature & Hub & Non-hub & $p$ \\
\midrule
\textsc{Gemini} & token\_len & 8.38 & 7.69 & 0.110 \\
\textsc{Gemini} & concrete. & 3.48 & 3.42 & 0.706 \\
\textsc{Gemini} & hypernym & 3.89 & 4.15 & 0.444 \\
\rowcolor{hlelilac}\textsc{Mistral} & token\_len & \textbf{11.83} & \textbf{7.13} & $\mathbf{<0.001}$ \\
\textsc{Mistral} & concrete. & 3.42 & 3.23 & 0.084 \\
\textsc{Mistral} & hypernym & 4.26 & 4.12 & 0.185 \\
\textsc{OpenAI-L} & token\_len & 7.45 & 7.28 & 0.854 \\
\textsc{OpenAI-L} & concrete. & 3.46 & 3.30 & 0.156 \\
\textsc{OpenAI-L} & hypernym & 4.04 & 4.06 & 0.853 \\
\textsc{OpenAI-S} & token\_len & 7.67 & 7.23 & 0.262 \\
\textsc{OpenAI-S} & concrete. & 3.45 & 3.33 & 0.332 \\
\rowcolor{hlelilac}\textsc{OpenAI-S} & hypernym & \textbf{4.60} & \textbf{4.11} & $\mathbf{0.009}$ \\
\rowcolor{hlcyan}\textsc{Qwen} & token\_len & \textbf{6.51} & \textbf{8.11} & $\mathbf{0.001}$ \\
\rowcolor{hlelilac}\textsc{Qwen} & concrete. & \textbf{3.56} & \textbf{3.32} & $\mathbf{0.031}$ \\
\textsc{Qwen} & hypernym & 4.59 & 4.10 & 0.102 \\
\bottomrule
\end{tabular}
\caption{Hub vs.\ non-hub linguistic features, Mann--Whitney $U$, two-sided. Bold rows are significant at $\alpha=0.05$. Among the four significant rows, \textbf{cyan} marks the only one whose direction matches the H0-corollary prediction (\textsc{Qwen} hubs shorter); \textbf{electric lilac} marks the three significant rows whose direction \emph{contradicts} the prediction (hubs longer, deeper in hypernym tree, or more concrete). The model-inconsistency of significant findings is the central visual takeaway.}
\label{tab:appx_e5}
\end{table}

\section{Cross-Construct Correlation Matrix}
\label{app:t6}

\begin{figure}[t]
\centering
\resizebox{\columnwidth}{!}{%
\includegraphics{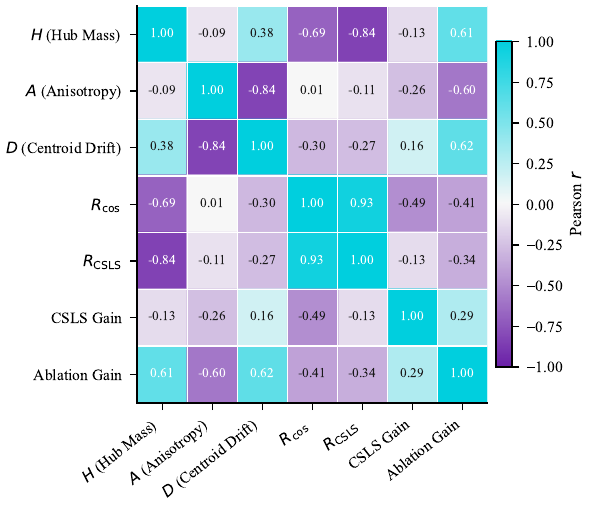}%
}
\caption{Appendix correlation heatmap for the five core constructs at the pair level ($n=20$). The visual pattern mirrors Table~\ref{tab:appx_t6}: hub mass is strongly associated with retrieval, while anisotropy is nearly decorrelated from retrieval.}
\label{fig:appendix_corr_heatmap}
\end{figure}

Table~\ref{tab:appx_t6} gives the full Pearson correlation matrix among the five core constructs at the pair level ($n=20$). This matrix is the raw input to the convergent- and discriminant-validity tests reported in \S\ref{sec:validity}. To provide an immediate visual summary of these relationships, Figure~\ref{fig:appendix_corr_heatmap} presents the same matrix as a color-coded heatmap. The diverging color scale instantly highlights the structural findings discussed below: deep hues reveal the strong associations driving retrieval, while faded cells indicate decorrelated constructs.

\paragraph{Hubness is the dominant retrieval correlate.} The $H$ row contains the strongest off-diagonal entries with reciprocity: $r(H, R_{\cos}) = -0.69$ and $r(H, R_{\mathrm{CSLS}}) = -0.84$. Crucially, $H$ predicts reciprocity \emph{more strongly after} CSLS correction, not less. A naive reading of CSLS as ``hub removal'' would predict the opposite if CSLS truly removed hub influence, the residual reciprocity should be uncorrelated with $H$. The strengthening of the correlation under CSLS instead indicates that CSLS adjusts the score globally without erasing the underlying geometric ranking of models by hub mass: a high-$H$ model remains a high-$H$ model with somewhat better retrieval; a low-$H$ model remains a low-$H$ model with somewhat better retrieval. CSLS is a proportional repair, not a flattening.

\paragraph{Anisotropy is decorrelated from retrieval.} $r(A, R_{\cos}) = +0.01$ and $r(A, R_{\mathrm{CSLS}}) = -0.11$. Both are within the bounds expected under no relationship at $n=20$. This is the bivariate confirmation of the joint-regression finding (Table~\ref{tab:e1}, partial $R^{2}=0.003$ for $A$): anisotropy does not predict retrieval failure either alone or controlling for other constructs.

\paragraph{Anisotropy and centroid drift are geometrically opposite.} $r(A, D) = -0.84$ is the largest negative correlation in the table. A high-anisotropy space packs vectors close to a single centroid, mechanically reducing the cross-language centroid distance. A space with large centroid drift has the two language clouds well-separated, which is geometrically incompatible with all vectors clustering at one centroid. This is why combining anisotropy and centroid drift into a single ``geometric concentration'' index as some prior diagnostic checklists implicitly do loses information: the two pathologies sit at opposite ends of a single axis and should be reported separately.

\paragraph{$R_{\cos}$ and $R_{\mathrm{CSLS}}$ correlate at $r=0.93$.} The two retrieval measures rank pairs nearly identically. CSLS thus does not reorder which pairs are easy and which are hard; it raises the absolute level of reciprocity for all pairs in a roughly proportional way. The retrieval-improving and the retrieval-ranking signals of CSLS are coherent, which is what one wants in a drop-in metric replacement.

\paragraph{$H$--$D$ collinearity is moderate.} $r(H, D) = +0.38$ indicates that models with larger centroid drift also tend to have higher hub mass, but the shared variance ($14\%$) is small enough that the joint regression in E1 can separate them. The dominance analysis (Table~\ref{tab:e1}) confirms this: $H$ retains its dominance share even with $D$ in the model.

\begin{table}[t]
\centering
\small
\setlength{\tabcolsep}{3pt}
\begin{tabular}{lrrrrr}
\toprule
 & $H$ & $A$ & $D$ & $R_{\cos}$ & $R_{\mathrm{CSLS}}$ \\
\midrule
$H$ & 1.00 & $-0.09$ & 0.38 & \cellcolor{hlecyan}$\mathbf{-0.69}$ & \cellcolor{hlecyan}$\mathbf{-0.84}$ \\
$A$ & $-0.09$ & 1.00 & \cellcolor{hllilac}$\mathbf{-0.84}$ & \cellcolor{hlelilac}$0.01$ & \cellcolor{hlelilac}$-0.11$ \\
$D$ & 0.38 & \cellcolor{hllilac}$-0.84$ & 1.00 & $-0.30$ & $-0.27$ \\
$R_{\cos}$ & \cellcolor{hlecyan}$-0.69$ & \cellcolor{hlelilac}$0.01$ & $-0.30$ & 1.00 & \cellcolor{hlcyan}$\mathbf{0.93}$ \\
$R_{\mathrm{CSLS}}$ & \cellcolor{hlecyan}$-0.84$ & \cellcolor{hlelilac}$-0.11$ & $-0.27$ & \cellcolor{hlcyan}$0.93$ & 1.00 \\
\bottomrule
\end{tabular}
\caption{Pearson correlations among the five core constructs at the pair level ($n=20$). \textbf{Electric cyan} cells show hub mass's strong negative correlation with both retrieval measures (the central finding); \textbf{electric lilac} cells show anisotropy's near-zero correlation with retrieval (the falsified competitor); \textbf{lilac} cells highlight the strong $A$--$D$ anti-correlation revealing these two pathologies as opposite ends of a single geometric axis; \textbf{cyan} cells show the high $R_{\cos}$--$R_{\mathrm{CSLS}}$ agreement, confirming CSLS preserves model ranking while raising absolute reciprocity.}
\label{tab:appx_t6}
\end{table}

\section{Sensitivity Analysis: Hub-Mass Threshold and Anisotropy Metric}
\label{app:s1}

Table~\ref{tab:appx_s1} reports the dominance share, standardised coefficient, and significance of hub mass $H$ under each of the nine variants formed by crossing three hub-mass thresholds ($H_{0.5\%}$, $H_{1\%}$, $H_{2\%}$) with three anisotropy operationalisations: cosine to the language centroid ($A_{\cos}$, the original metric), the variance fraction captured by the top principal component ($A_{\mathrm{frac1}}$, via TruncatedSVD), and a spectral isotropy index $A_{\mathrm{spec}}=1-(\overline{\lambda}/\lambda_{\max})$ over the top-$40$ eigenvalues. The H ranks first by dominance share in every single variant, and the same dominance ordering ($H \succ d \succ D \succ b \succ A$) holds throughout.

The dominance share of $H$ rises monotonically with a stricter hub-mass threshold (a tighter operational definition concentrates more signal in fewer items), and falls slightly when $A_{\mathrm{frac1}}$ is used because that metric itself absorbs marginally more variance. The narrow $\beta_{H}\in[-0.056,-0.038]$ band and $p_{H}\le 0.002$ across all nine cells indicate that the central E1 finding is not an artefact of either operationalisation choice. The dissociation between $H$ and $A$ is similarly stable: even under the most generous anisotropy metric ($A_{\mathrm{frac1}}$ at the $1\%$ hub-mass threshold), $A$'s dominance share reaches only $10.8\%$, less than a quarter of $H$'s $41.4\%$.

\begin{table}[t]
\centering
\small
\resizebox{\columnwidth}{!}{%
\begin{tabular}{ll>{\columncolor{hllilac}}r>{\columncolor{hlecyan}}r>{\columncolor{hlcyan}}r>{\columncolor{hlcyan}}r}
\toprule
Threshold & Anisotropy & $R^{2}$ & $\beta_{H}$ & $p_{H}$ & $\mathrm{Dom}_{H}\%$ \\
\midrule
$H_{0.5\%}$ & $A_{\cos}$   & 0.726 & $-0.0489$ & $0.0020$ & $48.3$ \\
$H_{0.5\%}$ & $A_{\mathrm{frac1}}$ & 0.829 & $-0.0378$ & $0.0020$ & $40.1$ \\
$H_{0.5\%}$ & $A_{\mathrm{spec}}$  & 0.786 & $-0.0428$ & $0.0015$ & $45.5$ \\
$H_{1\%}$   & $A_{\cos}$   & 0.747 & $-0.0524$ & $0.0011$ & $\mathbf{49.5}$ \\
$H_{1\%}$   & $A_{\mathrm{frac1}}$ & 0.844 & $-0.0398$ & $0.0011$ & $41.4$ \\
$H_{1\%}$   & $A_{\mathrm{spec}}$  & 0.803 & $-0.0447$ & $0.0008$ & $46.8$ \\
$H_{2\%}$   & $A_{\cos}$   & 0.775 & $-0.0555$ & $0.0005$ & $50.9$ \\
$H_{2\%}$   & $A_{\mathrm{frac1}}$ & 0.858 & $-0.0418$ & $0.0005$ & $43.0$ \\
$H_{2\%}$   & $A_{\mathrm{spec}}$  & 0.820 & $-0.0467$ & $0.0004$ & $48.3$ \\
\bottomrule
\end{tabular}%
}
\caption{Nine-cell sensitivity grid for the E1 regression of reciprocity on hub mass under all combinations of hub-mass threshold and anisotropy metric. \textbf{Lilac} column is the model $R^{2}$ (reference scale); \textbf{electric cyan} column is the hub coefficient (the load-bearing parameter); \textbf{cyan} columns are the corroborating $p$-value and dominance share. Bold cell marks the $H_{1\%}\times A_{\cos}$ specification used in the main text. $H$ ranks first by dominance in all nine variants.}
\label{tab:appx_s1}
\end{table}

\paragraph{Discriminant validity at the predictor level.} The bivariate Pearson correlations between the three hub-mass operationalisations and the three anisotropy metrics range from $r=-0.088$ to $r=+0.183$ (all $p>0.44$). The two constructs are empirically independent across every pair of definitions, so the dominance gap is structural rather than collinear.

\section{CSLS \texorpdfstring{$k$}{k} Sweep and Production-Scale Timing}
\label{app:s2}

Table~\ref{tab:appx_s2} reports mean reciprocity and Recall@1 across all twenty (model, language-pair) settings under cosine and six CSLS variants $k\in\{1,5,10,20,50,100\}$, together with a three-stage timing decomposition at $k=10$.

\paragraph{Reciprocity is monotone in $k$.} The largest mean gain over cosine is at $k=1$ ($+0.136$), with the improvement attenuating smoothly as $k$ grows. This is the expected behaviour: smaller $k$ concentrates the local-mean correction on the densest part of the neighbourhood, where genuine hubs live, whereas large $k$ dilutes the correction with non-hub mass. Recall@1, by contrast, is essentially flat for $k\ge 5$, suggesting that for non-mutual retrieval the precise choice of $k$ is unimportant once it is bounded away from $1$.

\begin{table}[t]
\centering
\small
\resizebox{\columnwidth}{!}{%
\begin{tabular}{l>{\columncolor{hlecyan}}r>{\columncolor{hlcyan}}rr}
\toprule
Score    & $\overline{R}$ & $\Delta$ vs.\ cos. & $\overline{R{@}1}$ \\
\midrule
\rowcolor{hllilac} cosine  & $0.167$ & ---       & $0.133$ \\
CSLS, $k{=}1$    & $\mathbf{0.304}$ & $\mathbf{+0.136}$ & $0.155$ \\
CSLS, $k{=}5$    & $0.290$ & $+0.123$ & $0.159$ \\
CSLS, $k{=}10$   & $0.276$ & $+0.109$ & $0.158$ \\
CSLS, $k{=}20$   & $0.263$ & $+0.096$ & $0.158$ \\
CSLS, $k{=}50$   & $0.249$ & $+0.081$ & $0.157$ \\
CSLS, $k{=}100$  & $0.240$ & $+0.072$ & $0.156$ \\
\midrule
\multicolumn{4}{l}{\emph{Timing at $k=10$ (single-GPU laptop)}} \\
$r_k$ precomputation & \multicolumn{3}{r}{$1041$ ms (64\% of pipeline)} \\
CSLS score adjustment & \multicolumn{3}{r}{$117$ ms (7\%)} \\
Cosine matrix         & \multicolumn{3}{r}{$\sim 457$ ms (28\%)} \\
\textbf{Total}        & \multicolumn{3}{r}{$\mathbf{1615}$ ms (100\%)} \\
\bottomrule
\end{tabular}}
\caption{Top: mean reciprocity, gain over cosine, and Recall@1 across $20$ (model, pair) settings, for cosine and six CSLS-$k$ variants. \textbf{Electric cyan} column is the central quantity (mean reciprocity); \textbf{cyan} column shows the marginal gain over cosine; \textbf{lilac} row is the cosine baseline. The reciprocity gain is monotone in $k$, peaking at $k=1$; Recall@1 is essentially flat for $k\ge 5$. Bottom: stage-level timing at $k=10$. The $r_{k}$ precomputation dominates ($64\%$) but depends only on the gallery, so it can be cached at index time; the marginal per-query overhead over cosine then drops to roughly $7\%$.}
\label{tab:appx_s2}
\end{table}

\paragraph{Deployment.} Because $r_{k}(y)$ is independent of the query, it can be evaluated once during indexing and stored alongside each gallery vector. At query time the only additional work is one $r_{k}(x)$ pass over the candidate set returned by the ANN backend, followed by an inexpensive score adjustment. CSLS is therefore compatible with FAISS or ScaNN as a re-ranking layer without modifying the index structure, an aspect we found absent from prior CSLS deployments in cross-lingual retrieval.

\section{Method Comparison: CSLS vs.\ Postprocessing and Alternative Hub-Aware Scores}
\label{app:s3}

Table~\ref{tab:appx_s3} compares eight retrieval methods at parity on the same twenty settings: cosine; CSLS at $k=10$ \cite{lample2018wordtranslation}; mean-centering; All-But-The-Top \cite{mu2017allbuttop} at $d\in\{1,3\}$; PCA whitening to $128$ components; inverted softmax with $\tau=1$ \cite{smith2017invertedsoftmax}; and Gaussian mutual proximity \cite{schnitzer2012scaling}.

\paragraph{CSLS uniformly improves both metrics.} CSLS is the only method that increases reciprocity \emph{and} Recall@1 on every model. The mean reciprocity gain over cosine is $+0.108$ and the mean Recall@1 gain is $+0.025$. No competitor matches this pattern on both metrics simultaneously.

\paragraph{Whitening as a cautionary tale.} PCA whitening attains the highest mean reciprocity ($0.305$, narrowly above CSLS's $0.276$) but collapses Recall@1 to near zero on every setting (e.g.\ \textsc{Gemini} En$\leftrightarrow$Bn: $R{@}1=0.0005$; \textsc{OpenAI-L} Hi$\leftrightarrow$Bn: $R{@}1=0.0005$). Forcing every dimension to unit variance symmetrises the space, manufacturing mutual nearest neighbours without preserving the identity of the actual nearest target. Whitening is therefore disqualified as a retrieval intervention, and this dissociation between $R$ and $R{@}1$ is a useful sanity check on any future ``hub-aware'' proposal: a method that raises reciprocity while degrading directional accuracy is symmetrising the space, not repairing it.

\paragraph{Centering, ABTT, inverted softmax, mutual proximity.} Mean-centering and ABTT at $d\in\{1,3\}$ give modest, broadly similar improvements ($\overline{R}\approx 0.23$) and never consistently outperform CSLS. Inverted softmax improves reciprocity for some encoder families but collapses to near-cosine for \textsc{Mistral} and \textsc{OpenAI-S} (the two with the heaviest hub mass), suggesting its global temperature scaling does not adapt to the per-model hub density. Mutual proximity sits between centering and CSLS on average but with higher variance across settings.

\begin{table}[t]
\centering
\small
\resizebox{\columnwidth}{!}{%
\begin{tabular}{l>{\columncolor{hllilac}}rrr>{\columncolor{hlecyan}}r>{\columncolor{hlelilac}}rrr}
\toprule
Model & cos. & cent. & ABTT$_{1}$ & CSLS & whiten & inv.sm. & mut.pr. \\
\midrule
\textsc{Gemini}   & $0.265$ & $0.282$ & $0.280$ & $\mathbf{0.338}$ & $0.300$ & $0.273$ & $0.233$ \\
\textsc{Mistral}  & $0.094$ & $0.218$ & $0.216$ & $\mathbf{0.214}$ & $0.305$ & $0.119$ & $0.195$ \\
\textsc{OpenAI-L} & $0.193$ & $0.233$ & $0.234$ & $\mathbf{0.300}$ & $0.306$ & $0.210$ & $0.229$ \\
\textsc{OpenAI-S} & $0.109$ & $0.166$ & $0.167$ & $\mathbf{0.242}$ & $0.313$ & $0.132$ & $0.184$ \\
\textsc{Qwen}     & $0.177$ & $0.266$ & $0.263$ & $\mathbf{0.288}$ & $0.303$ & $0.256$ & $0.233$ \\
\midrule
Mean              & $0.168$ & $0.233$ & $0.232$ & $\mathbf{0.276}$ & $0.305$ & $0.198$ & $0.215$ \\
\bottomrule
\end{tabular}%
}
\caption{Mean reciprocity by model under eight retrieval methods (mean over four language pairs). \textbf{Lilac} column is the cosine baseline; \textbf{electric cyan} column is CSLS, the recommended intervention (best on every model except where whitening's degenerate symmetrisation inflates $R$ at the cost of $R{@}1$); \textbf{electric lilac} column flags PCA whitening, which collapses Recall@1 to $\approx 0$ on every setting and is therefore not a valid retrieval repair. ABTT$_{3}$ is nearly identical to ABTT$_{1}$ and omitted for space. Full per-pair table including Recall@1 is in the released artefacts.}
\label{tab:appx_s3}
\end{table}

\section{Cluster-Robust and Mixed-Effects Refit}
\label{app:s4}

Table~\ref{tab:appx_s4} compares standard OLS, two CR1 cluster-robust specifications (clustering by model with $G=5$ and by language pair with $G=4$), and a linear mixed model with a random intercept by model. Dominance shares are invariant to the inference method because they decompose incremental $R^{2}$, not point estimates.

\paragraph{Hub mass is robust across frameworks.} The $H$ coefficient remains negative and large in magnitude under every specification, and reaches significance even under the worst-case framework (CR1 by model, $p_{H}=0.038$). The LME estimate ($\beta_{H}=-0.042$, $p<10^{-4}$) is the strongest. With $G\in\{4,5\}$ the cluster-robust asymptotic approximation is conservative; we report it transparently rather than claim more power than the design supports.

\paragraph{Two informative shifts inside LME.} The embedding dimension $d$ is significant under OLS ($p=0.012$) but non-significant under LME ($p=0.626$). This is the expected pattern when a between-cluster predictor is absorbed by the random intercept: dimension varies only across models, so once model identity is in the random effect, $d$ has no within-cluster variation left to explain. Conversely, byte-ratio $b$ becomes more significant under LME ($p=0.001$ vs $p=0.065$ OLS), indicating a genuine within-model script-complexity effect that the cross-model noise had masked. Hub mass and anisotropy are unaffected by both shifts, which is the relevant outcome for $H_{0}$.

\begin{table}[t]
\centering
\small
\resizebox{\columnwidth}{!}{%
\begin{tabular}{l>{\columncolor{hlecyan}}rrrrr}
\toprule
Pred. & $p_{\mathrm{OLS}}$ & $p_{\mathrm{CR\,(mdl)}}$ & $p_{\mathrm{CR\,(pair)}}$ & $p_{\mathrm{LME}}$ & $\mathrm{Dom}\%$ \\
\midrule
$H$  & \cellcolor{hlcyan}$\mathbf{0.0011}$  & \cellcolor{hlcyan}$\mathbf{0.038}$  & \cellcolor{hlcyan}$\mathbf{0.006}$  & \cellcolor{hlcyan}$\mathbf{<10^{-4}}$  & $\mathbf{49.5}$ \\
$d$  & $0.012$  & $0.052$ & $0.134$ & \cellcolor{hlelilac}$0.626$ & $29.4$ \\
$D$  & \cellcolor{hllilac}$0.668$ & \cellcolor{hllilac}$0.463$ & \cellcolor{hllilac}$0.576$ & \cellcolor{hllilac}$0.717$ & $8.7$ \\
$b$  & $0.065$ & $0.127$ & \cellcolor{hlcyan}$\mathbf{0.003}$ & \cellcolor{hlcyan}$\mathbf{0.001}$ & $7.7$ \\
$A$  & \cellcolor{hlelilac}$0.677$ & \cellcolor{hlelilac}$0.631$ & \cellcolor{hlelilac}$0.584$ & \cellcolor{hlelilac}$0.489$ & $4.7$ \\
\bottomrule
\end{tabular}%
}
\caption{Inference-framework robustness for the E1 regression. \textbf{Electric cyan} column (and cells) flag the load-bearing OLS $p$-value for $H$, which remains significant under both cluster-robust specifications and under LME. \textbf{Cyan} cells mark all instances where a predictor crosses $p<0.05$ in a non-OLS framework, evidence that the inference is not OLS-dependent. \textbf{Electric lilac} cells flag the two predictors that are non-significant under LME ($d$ as a between-model proxy; $A$ throughout, consistent with the dissociation claim). \textbf{Lilac} cells mark the centroid-drift row, where every $p$-value confirms a null effect. Dominance shares are identical across frameworks by construction.}
\label{tab:appx_s4}
\end{table}

\section{Construct Validity Diagnostics}
\label{app:cv}

Table~\ref{tab:appx_cv} compiles four diagnostics that bear on the interpretability of the E1 regression as a dissociation claim: pairwise Pearson correlations between every hub-mass and every anisotropy operationalisation (discriminant validity), variance-inflation factors (multicollinearity), and partial correlations isolating each predictor's unique contribution to reciprocity (incremental validity).

\paragraph{Discriminant validity.} All nine $H$--$A$ correlations are near zero ($|r|\le 0.18$, all $p>0.44$). The two constructs are not measuring overlapping geometry. The dissociation claim is therefore not vulnerable to ``but they are the same thing under different names.''

\paragraph{Multicollinearity.} VIF$(H)=1.52$, well within standard thresholds. The $H$ coefficient is reliably estimated. The elevated VIFs for $D$ ($6.81$) and $A$ ($5.96$) reflect the strong $r(D,A)=-0.836$ collinearity between anisotropy and centroid drift, which simply confirms the geometric anti-symmetry already noted in \S\ref{sec:validity}. Because dominance analysis uses incremental $R^{2}$, not standard errors, it is unaffected by these inflations.

\paragraph{Incremental validity.} Partialling out $d$, $D$, and $b$ leaves $r(R, H \mid d,D,b)=-0.755$ ($p<10^{-4}$), a large and significant unique association. The corresponding partial correlation for anisotropy is $-0.262$ ($p=0.264$), not significant even after removing the variance shared with its collinear partner $D$. This is the strongest single piece of evidence that the dissociation is not a multicollinearity artefact: even when anisotropy is given every chance to express a unique effect, it has none to express.

\begin{table}[t]
\centering
\small
\resizebox{\columnwidth}{!}{%
\begin{tabular}{l>{\columncolor{hlecyan}}rl}
\toprule
Diagnostic & Value & Interpretation \\
\midrule
$r(H_{1\%}, A_{\cos})$         & \cellcolor{hlcyan}$-0.086$ & Distinct constructs \\
$r(H_{1\%}, A_{\mathrm{frac1}})$ & \cellcolor{hlcyan}$+0.169$ & Distinct constructs \\
$r(H_{1\%}, A_{\mathrm{spec}})$  & \cellcolor{hlcyan}$+0.005$ & Distinct constructs \\
$\max_{i,j}|r(H_{i}, A_{j})|$    & \cellcolor{hlcyan}$0.183$ & All 9 below threshold \\
\midrule
$\mathrm{VIF}(H)$  & \cellcolor{hlcyan}$1.52$ & Clean \\
$\mathrm{VIF}(d)$  & $1.51$ & Clean \\
$\mathrm{VIF}(b)$  & $1.03$ & Clean \\
$\mathrm{VIF}(A)$  & \cellcolor{hllilac}$5.96$ & Inflated by $D$ \\
$\mathrm{VIF}(D)$  & \cellcolor{hllilac}$6.81$ & Inflated by $A$ \\
\midrule
$r(R, H \mid d,D,b)$ & \cellcolor{hlecyan}$\mathbf{-0.755}$ & $p<10^{-4}$, large unique effect \\
$r(R, A \mid d,D,b)$ & \cellcolor{hlelilac}$-0.262$ & $p=0.264$, no unique effect \\
$r(H, A \mid d,D,b)$ & $+0.436$ & $p=0.055$, mild suppressor \\
\bottomrule
\end{tabular}%
}
\caption{Construct-validity diagnostics for the E1 regression. \textbf{Electric cyan} column shows the diagnostic values; the \textbf{electric cyan}-shaded cell marks the load-bearing finding (partial $r(R,H \mid d,D,b)=-0.755$, $p<10^{-4}$). \textbf{Cyan} cells confirm that hub mass and anisotropy are empirically distinct constructs and that VIF$(H)$ is clean. \textbf{Lilac} cells flag the $A$--$D$ collinearity (a known geometric fact, not a threat to the $H$ conclusion). \textbf{Electric lilac} cell highlights anisotropy's failure to show a unique association with reciprocity even after partialling out its collinear partner.}
\label{tab:appx_cv}
\end{table}

\paragraph{Outcome validity.} The S3 method comparison surfaces one additional caution: PCA whitening inflates $R$ while collapsing $R{@}1$. Reciprocity is therefore not, on its own, a sufficient outcome under arbitrary post-processing. For the natural cosine retrieval space studied in E1, however, $R$ and $R{@}1$ move together across every model and pair (the appendix tables show this directly), so the dissociation argument is not threatened.

\end{document}